\documentclass{article}

\usepackage{microtype}
\usepackage{graphicx}
\usepackage{subcaption}
\usepackage{booktabs} 

\usepackage{hyperref}

\usepackage[preprint]{icml2026}

\usepackage{amsmath}
\usepackage{amssymb}
\usepackage{mathtools}
\usepackage{amsthm}

\usepackage{multirow}
\usepackage{multicol}
\usepackage{booktabs}
\usepackage{xcolor}
\usepackage{colortbl}
\definecolor{citecolor}{RGB}{34,139,34}
\definecolor{demphcolor}{gray}{.5}
\definecolor{Graylight}{gray}{0.9}
\newcommand{\demph}[1]{\textcolor{demphcolor}{#1}}

\newlength\savewidth\newcommand\shline{\noalign{\global\savewidth\arrayrulewidth
  \global\arrayrulewidth 1pt}\hline\noalign{\global\arrayrulewidth\savewidth}}
\newcommand{\tablestyle}[2]{\setlength{\tabcolsep}{#1}\renewcommand{\arraystretch}{#2}\centering\footnotesize}
\renewcommand{\paragraph}[1]{\vspace{1.25mm}\noindent\textbf{#1}}

\usepackage{pifont}

\usepackage[capitalize,noabbrev]{cleveref}

\theoremstyle{plain}

\theoremstyle{definition}

\theoremstyle{remark}

\usepackage[textsize=tiny]{todonotes}

\icmltitlerunning{VTok: A Unified Video Tokenizer with Decoupled Spatial-Temporal Latents}

\begin{document}

\twocolumn[

  \icmltitle{VTok: A Unified Video Tokenizer with Decoupled Spatial-Temporal Latents}



  \icmlsetsymbol{equal}{*}

  \begin{icmlauthorlist}
    \icmlauthor{Feng Wang}{seed,jhu}
    \icmlauthor{Yichun Shi}{seed}
    \icmlauthor{Ceyuan Yang}{seed}
    \icmlauthor{Qiushan Guo}{seed}
    \icmlauthor{Jingxiang Sun}{seed}
    \icmlauthor{Alan Yuille}{jhu}
    \icmlauthor{Peng Wang}{seed}

  \end{icmlauthorlist}

  \icmlaffiliation{seed}{Bytedance Seed}
  \icmlaffiliation{jhu}{Johns Hopkins University}

  \icmlcorrespondingauthor{Peng Wang}{peng.wang@bytedance.com}

  \icmlkeywords{Machine Learning, ICML}

  \vskip 0.3in
]



\printAffiliationsAndNotice{}  

\begin{abstract}

This work presents VTok, a unified video tokenization framework that can be used for both generation and understanding tasks. Unlike the leading vision–language systems that tokenize videos through a naïve frame-sampling strategy, we propose to decouple the spatial and temporal representations of videos by retaining the spatial features of a single key frame while encoding each subsequent frame into a single residual token, achieving compact yet expressive video tokenization. Our experiments suggest that VTok effectively reduces the complexity of video representation from the product of frame count and per-frame token count to their sum, while the residual tokens sufficiently capture viewpoint and motion changes relative to the key frame. Extensive evaluations demonstrate the efficacy and efficiency of VTok: it achieves notably higher performance on a range of video understanding and text-to-video generation benchmarks compared with baselines using naïve tokenization, all with shorter token sequences per video (e.g., 3.4\% higher accuracy on our TV-Align benchmark and 1.9\% higher VBench score). Remarkably, VTok produces more coherent motion and stronger guidance following in text-to-video generation, owing to its more consistent temporal encoding. We hope VTok can serve as a standardized video tokenization paradigm for future research in video understanding and generation. Webpage:  \url{https://wangf3014.github.io/VTok_page}.

\end{abstract}

\begin{figure}[t]
    \centering
    \includegraphics[width=\linewidth]{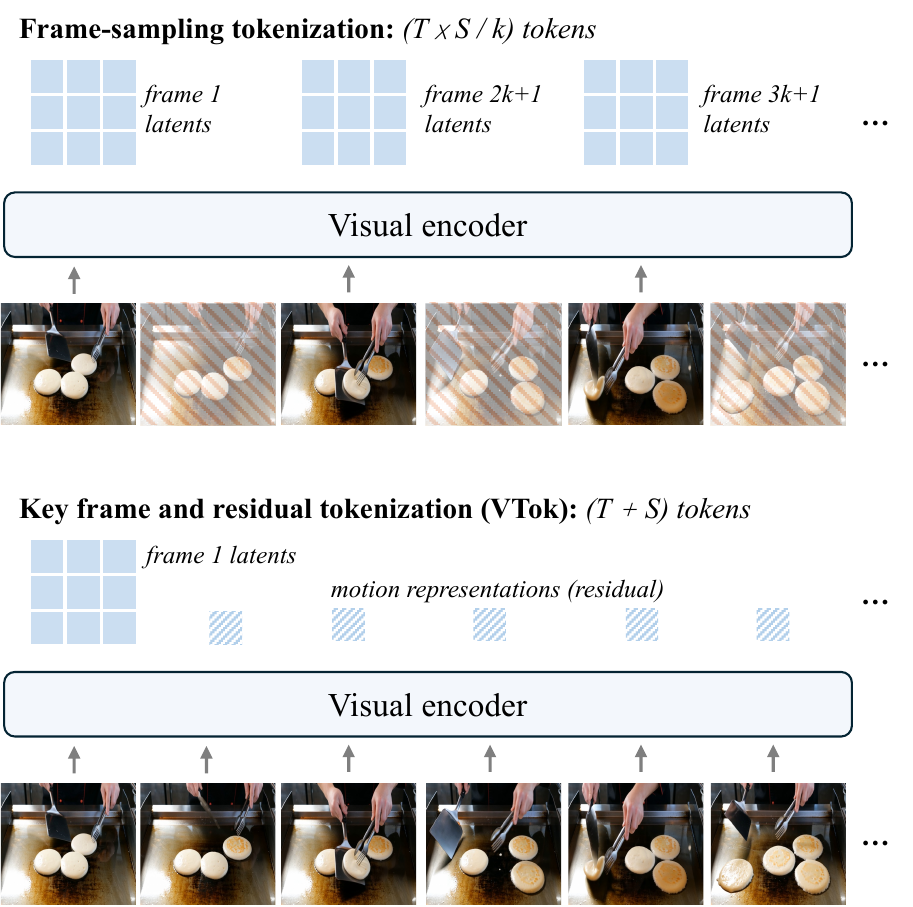}
    \caption{\textit{\textbf{Video tokenization strategies.}} Leading VLMs such as Qwen2.5-VL~\cite{qwen25-vl} still rely on a naïve frame-sampling approach to process videos. We contend that this strategy introduces excessive redundancy in spatial information while omitting temporal details. We find that video tokenization can be decomposed into spatial and temporal components, which preserves spatiotemporal information while minimizing the overall token length.}
    \label{fig:token}
    \vspace{-0.4cm}
\end{figure}

\section{Introduction}

\begin{figure*}[t]
    \centering
    \includegraphics[width=\textwidth]{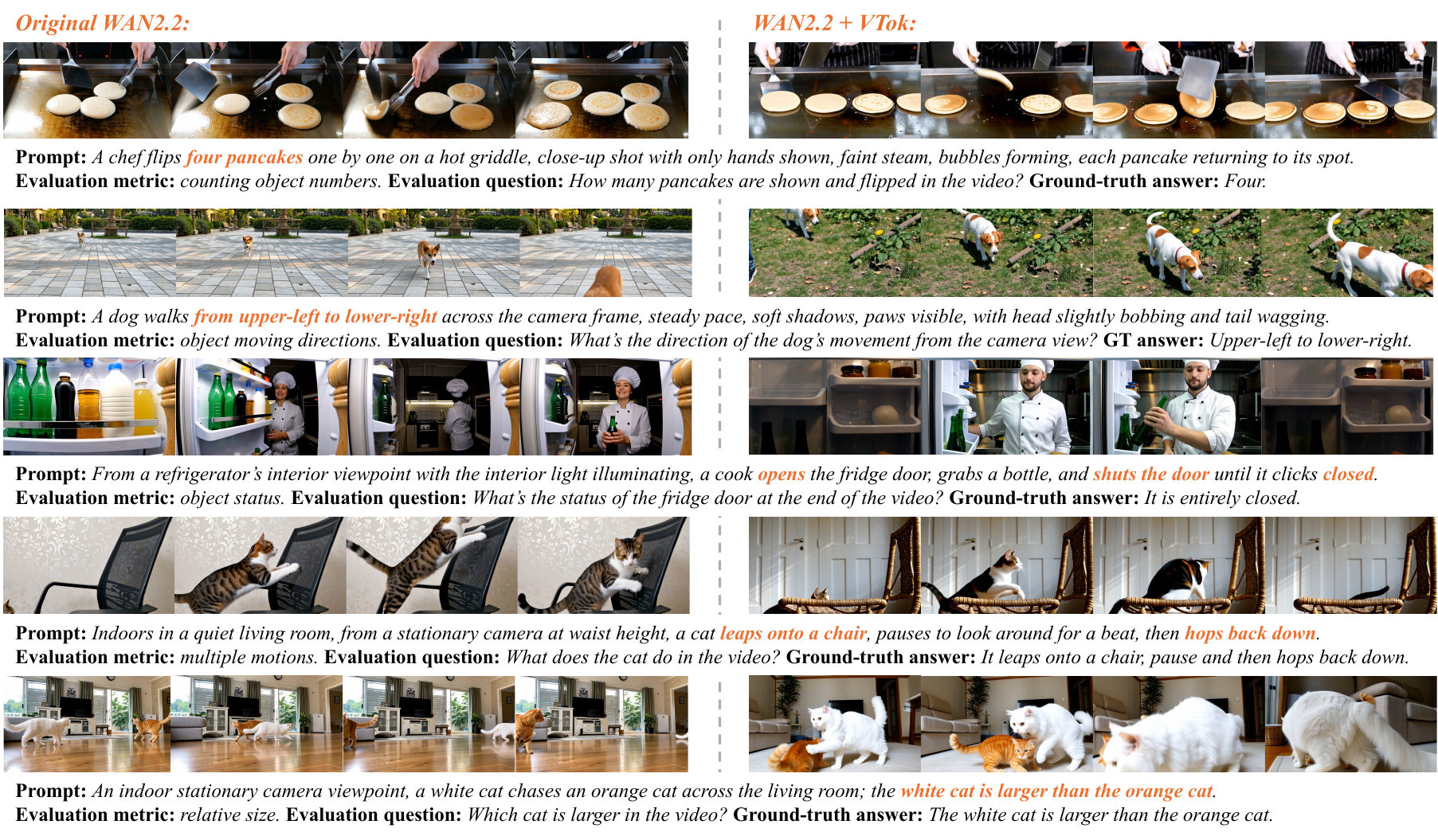}
    \vspace{-0.5cm}
    \caption{\textit{\textbf{Video generation examples.}} We showcase the video generation results of the state-of-the-art open-source model WAN2.2~\cite{wan} on our text-to-video alignment benchmark. By integrating VTok, the model demonstrates a more precise understanding of textual guidance, showing noticeably improved accuracy in following attributes such as object count, motion direction, position, and size.}
    \label{fig:examples}
    \vspace{-0.2cm}
\end{figure*}

A central bottleneck in scaling multimodal models to video is the interface between pixels and language—how to compress long, high‑resolution streams into tokens that an LLM can reason over. The dominant practice in recent video–language systems (\emph{e.g.}, Qwen2.5‑VL~\cite{qwen25-vl}, LLaVANext‑Video~\cite{llavanext-video}, BLIP-3-Video~\cite{blip-3-video}) is to first sample frames, encode each frame with a 2D image encoder, and then concatenate all visual tokens across frames into one long sequence that is fed to the language model. This design inherits strong spatial features but creates a hard trade‑off: either sample very sparsely to keep the token budget manageable—thereby erasing micro‑motions and transient events—or retain more frames and suffer a rapidly growing sequence length, which drives up quadratic attention cost and makes long‑range temporal credit assignment brittle. As a consequence, for understanding, models often struggle with fine‑grained temporal reasoning and event ordering~\cite{qwen25-vl,blip-3-video,llava-video,llavanext-video,internvideo,internvideo2,v-jepa}; for generation, textual guidance is often weakly grounded in motion, leading to timing inconsistencies and identity drift despite acceptable performance on coarse recognition~\cite{cogvideox,cosmos,hunyuanvideo,wan}.

In fact, given the strong prior of inter‑frame continuity and coherence, frames within a scene of moderate duration tend to maintain very similar spatial structure while changing smoothly in viewpoint, object pose, and location. This implies that frame‑sampling–based video tokenization introduces substantial redundancy on the spatial side while excessively discarding subtle temporal variations. This imbalance motivates us to develop an efficient video tokenization method that streamlines the spatial structure of a video while preserving as much motion information as possible. Through a series of investigations across both video understanding and generation tasks, we surprisingly find that video tokenization can be explicitly decoupled into spatial and temporal components. Concretely, as shown in Figure~\ref{fig:token}, a video clip suffices to retain the full spatial information of the first frame (often the key frame), while encoding each subsequent frame as a single token that captures its temporal change relative to the key frame. Under this paradigm, the token budget for a clip reduces from $\mathcal{O}(T\times S)$ to $\mathcal{O}(T+S)$—with $S$ spatial tokens from the key frame and $T$ frames—substantially improving efficiency while preserving action and motion details.

\begin{figure*}[t]
    \centering
    \includegraphics[width=\linewidth]{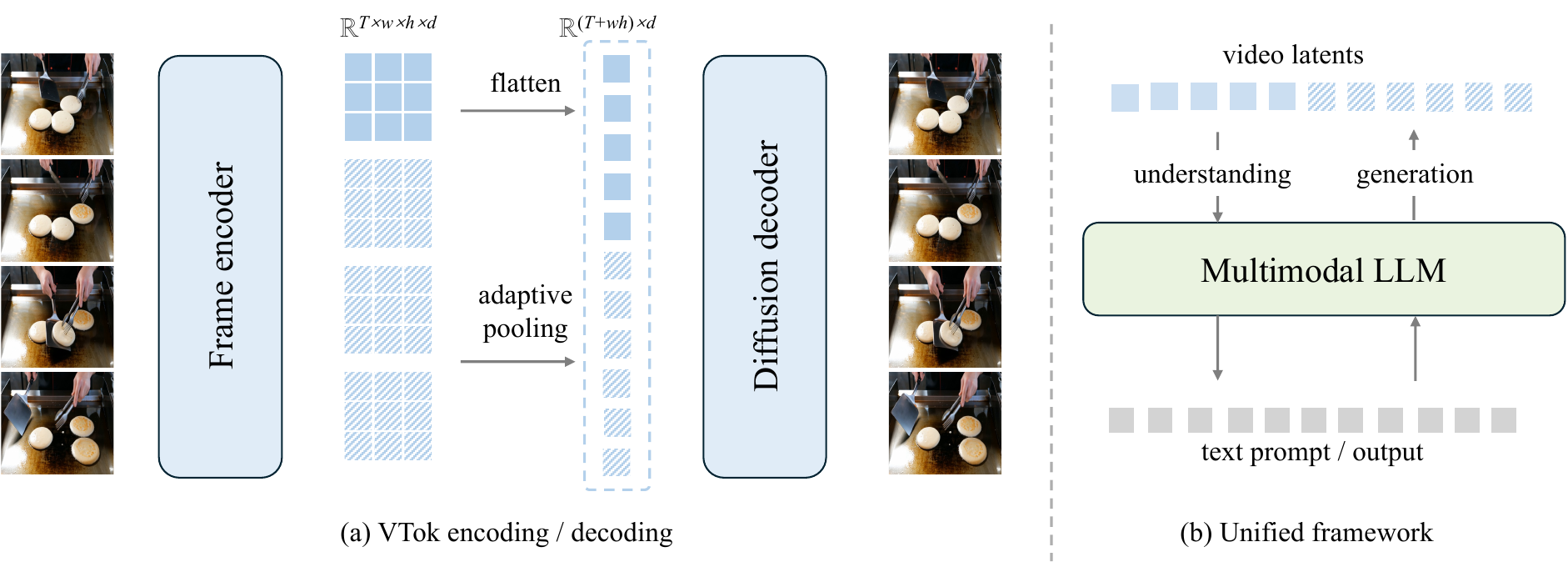}
    \caption{\textbf{\textit{Our unified framework of video understanding and text-to-video generation.}} The model integrates video understanding and generation within a single autoregressive multimodal large language model (MLLM). In the understanding branch, videos are first tokenized into key-frame and residual tokens, aligned with the language space, and processed together with a textual prompt to produce semantic outputs such as captions or answers. In the generation branch, the MLLM samples visual tokens conditioned on text, following the same spatial–temporal format, and a diffusion transformer can decode these tokens back to a video.}
    \label{fig:framework}
    \vspace{-0.4cm}
\end{figure*}

We validate this new tokenization paradigm by integrating it into a unified framework that jointly learns from video-to-text representation and text-to-video generation tasks. As shown in Figure~\ref{fig:framework}, our model is built on a unified autoregressive Multimodal Large Language Model (MLLM) that shares the same tokenization and reasoning backbone across both understanding and generation. For the understanding branch, a video is encoded into a sequence of key-frame and residual tokens, which are projected into the same embedding space as language tokens. These visual tokens are then concatenated with an understanding-oriented textual prompt and processed by the MLLM to produce textual outputs such as captions, answers, or explanations. In the generation branch, the same MLLM takes a textual prompt as input and autoregressively samples visual tokens that conform to our key-frame-plus-residual format. The resulting token sequence is subsequently decoded by a pretrained diffusion transformer into the final video, ensuring both spatial fidelity and temporally consistent motion generation.

We evaluate our model by testifying its performance on video understanding benchmarks such as Video-MMMU~\cite{video-mmmu}, its generation quality under the VBench~\cite{vbench} standard, as well as our newly proposed Text-to-Video Guidance Alignment Benchmark (TV-Align), which highlights whether generated videos faithfully follow high-level semantic instructions in prompts—such as object count, direction, and relative position. Results show that our model achieves notably stronger generation performance, outperforming the base model by 1.92 and 4.33 points on the total and semantic metrics of VBench~\cite{vbench}, respectively. This demonstrates that our tokenization paradigm provides more precise spatiotemporal guidance, leading to scenes and motions that are both semantically grounded and visually coherent. On the TV-Align benchmark, our model significantly surpasses leading open-source systems such as HunyuanVideo~\cite{hunyuanvideo} and Wan2.2~\cite{wan}, validating the superiority of our text-to-video generation framework—in which the MLLM itself generates visual tokens autoregressively, rather than merely conditioning a diffusion transformer on textual embeddings. This design enhances cross-modal semantic encoding, thereby achieving stronger text–video alignment. On video understanding benchmarks, our model likewise demonstrates consistent and significant gains. A fine-tuned LLaVA-Next~\cite{llavanext-video} model with a LLaMA-3~\cite{llama3} text encoder achieves an average improvement of 2.3\% over the original base model across seven standard benchmarks, confirming that our unified training and tokenization strategy effectively benefits both understanding and generation tasks. We hope that our new video tokenizer and unified generative–understanding framework can provide valuable insights and foundations for future research on efficient and semantically aligned video–language modeling.

\section{Related Work}

\noindent \textbf{Image and video tokenization} A classical approach to visual tokenization is to compress visual information into a compact continuous~\cite{ldm,dit,sit,softvq,uvit,mdtv2,repa} or discrete~\cite{vqgan,maskgit,llamagen,mage,titok,unitoken} latent space through a variational autoencoder (VAE) framework, where an encoder–decoder structure jointly performs information compression and reconstruction. This paradigm was initially developed for 2D images and was later extended to videos, showing that temporal data can be tokenized and reconstructed in a similar manner~\cite{hunyuanvideo,wan,cosmos,cogvideox,opensora,opensora2,sparse-vidgen,elastictok,tokenlearner,videoldm,videopoet,jang2025efficient,vidtok,omnitokenizer,emu3,loong,magvitv2,atoken,gigatok,moviegen,landiff,owl1,lct,instructvideo,latte,ltx-video,animatediff,videocrafter2,vchitect}. By pairing such video tokenizers with large-scale autoregressive language models or diffusion models, recent works have achieved impressive progress in both image and video generation and understanding~\cite{lwm,vila-u,emu3,llavanext-video,llava-ov,blip3,blip-3-video,qwen25-vl,vitok,siglip2,videoprism,veggie,vimi}. An alternative line of work leverages pretrained visual encoders to convert visual inputs into latent representations, which are then used as conditioning signals for external models such as multimodal large language models (MLLMs) or diffusion transformers to de-tokenize the signal back to the visual domain~\cite{yang2025rethinking,larp,divot,hitvideo}. Unlike VAE-style tokenization, this strategy does not emphasize precise pixel-level reconstruction; instead, it focuses on maintaining high-level semantic alignment between visual and textual modalities, enabling efficient multimodal reasoning and instruction following. This work mainly focuses on the later. Among these video tokenization studies, several recent works are most aligned with our perspective. For example, OmniTokenizer~\cite{omnitokenizer} introduces a spatial–temporal dual-branch design, where a dedicated temporal branch provides additional motion information to complement key-frame encoding. Similarly, Video-LaVIT~\cite{video-lavit} incorporates a specialized motion encoder to obtain dual-stream video latents, motivated by the intuition that explicit motion cues can significantly enhance video understanding~\cite{videocomposer,shen2024decouple}.

\noindent \textbf{Foundational video models} The rise of video foundation models marks a shift toward large-scale, general-purpose video understanding systems~\cite{qwen25-vl,blip-3-video,siglip2,llava-ov,llava-video,llavanext-video,emu3}. These models aim to learn unified representations that capture both spatial semantics and temporal dynamics from massive video–text datasets. Early frameworks extended image–language pretraining~\cite{clip,sclip,siglip,flamingo,coca,align,basic,florence,blip} to videos by encoding sampled frames and aligning them with textual captions, enabling strong zero-shot reasoning across diverse tasks. Recent advances scale this idea further, training billions of parameters on diverse web videos to achieve robust cross-domain transfer in captioning, retrieval, question answering, and generation~\cite{hunyuanvideo,magvitv2,wan,opensora,opensora2,cosmos,emu3}. Despite impressive progress, most foundation models still rely on frame-based tokenization, which limits their efficiency and temporal fidelity, highlighting the need for more structured video representations like ours.

\section{Method}

\subsection{Unified Framework}

We build a unified model that jointly handles video understanding and video generation within a single multimodal framework. The architecture consists of two complementary branches: an understanding branch that maps videos to textual outputs, and a generation branch that synthesizes videos conditioned on textual prompts, both sharing the same autoregressive backbone and tokenization scheme. Formally, let a video be $X \in \mathbb{R}^{T \times H \times W \times 3}$ and a text prompt be $p$. Our framework couples a video encoder $\mathcal{T}_{\text{vid}}$ with a single autoregressive MLLM and a video decoder $\mathcal{D}_{\text{vid}}$. The same MLLM is used for both video-to-text understanding and text-to-video generation. We then denote the video tokenization pipeline as
\begin{equation}
\mathcal{T}_{\text{vid}}: \mathbb{R}^{T\times H\times W\times 3}\rightarrow \mathbb{R}^{L_v\times d_v},\quad 
V=\mathcal{T}_{\text{vid}}(X),
\end{equation}
which yields a sequence of $L_v$ visual tokens $V=\{v_1,\dots,v_{L_v}\}$.
Text is tokenized by a standard tokenizer:
\begin{equation}
\mathcal{T}_{\text{text}}: \mathcal{V}_{\text{str}}\rightarrow \{0,\dots,|\Sigma|-1\}^{L_t},\quad 
Y=\mathcal{T}_{\text{text}}(p),
\end{equation}
with embeddings $\phi_{\text{text}}:\mathbb{N}\rightarrow \mathbb{R}^{d}$ and a visual projection $\phi_{\text{vis}}:\mathbb{R}^{d_v}\rightarrow \mathbb{R}^{d}$.
The MLLM input is the concatenation
\begin{equation}
Z=\big[\,\phi_{\text{text}}(Y_{\text{in}})\ ;\ \phi_{\text{vis}}(V)\,\big] + P,
\end{equation}
where $P$ denotes positional or modality embeddings and the concatenation order depends on the branch. The MLLM defines a conditional distribution over token sequences:
\begin{equation}
f_\theta(Z) = \text{Transformer}_\theta(Z),
\end{equation}
\begin{equation}
p_\theta(s_i \mid s_{<i})=\text{softmax}\big(W\,f_\theta(Z)_i\big),
\end{equation}
where $s_i$ ranges over a unified vocabulary that includes both text and visual tokens after projection into the shared space.

In the understanding branch, given video $X$ and instruction prompt $p$, the input is arranged as
\begin{equation}
Z_{\text{under}}=\big[\,\phi_{\text{text}}(\mathcal{T}_{\text{text}}(p))\ ;\ \phi_{\text{vis}}(\mathcal{T}_{\text{vid}}(X))\,\big] + P.
\end{equation}
The model autoregressively predicts the target text sequence $Y^{\star}$ (caption, answer, rationale) using the standard language modeling loss:
\begin{equation}
\mathcal{L}_{\text{under}} = -\sum_{i=1}^{|Y^{\star}|}\log p_\theta\!\left(Y^{\star}_i \mid Y^{\star}_{<i},\, Z_{\text{under}}\right).
\end{equation}
In the generation branch, given prompt $p$, the model samples visual tokens $\hat V$ conditioned on text:
\begin{equation}
Z_{\text{gen}}=\big[\,\phi_{\text{text}}(\mathcal{T}_{\text{text}}(p))\,\big]+P,\qquad 
\hat V \sim p_\theta(\cdot \mid Z_{\text{gen}}).
\end{equation}
A pretrained video decoder $\mathcal{D}_{\text{vid}}$, a diffusion transformer learned in the latent space reconstructs the video:
\begin{equation}
\hat X=\mathcal{D}_{\text{vid}}(\hat V).
\end{equation}
Training uses a visual-token language modeling loss plus the decoder’s diffusion objective:
\begin{equation}
\mathcal{L}_{\text{visLM}} = -\sum_{j=1}^{L_v}\log p_\theta\!\left(V_j \mid V_{<j},\, Z_{\text{gen}}\right), 
\end{equation}
\begin{equation}
\mathcal{L}_{\text{dec}} =
\mathbb{E}_{t,\epsilon}\big[\|\epsilon - \epsilon_\psi(V_t, t)\|\big]
\end{equation}
The overall training objective combines both branches:
\begin{equation}
\mathcal{L} = \mathcal{L}_{\text{under}} + \lambda_{\text{visLM}}\mathcal{L}_{\text{visLM}} + \lambda_{\text{dec}}\mathcal{L}_{\text{dec}}.
\end{equation}
At inference, the same MLLM is used for:
(i) \textit{understanding}, by conditioning on $(p,X)$ and decoding text; and
(ii) \textit{generation}, by conditioning on $p$, sampling visual tokens, and decoding them to video via $\mathcal{D}_{\text{vid}}$.

\subsection{Spatial-Temporal Decoupled Tokenization}

Given the unified framework defined above, our goal is to design a video tokenizer $\mathcal{T}_{\text{vid}}$ that produces compact yet expressive representations by explicitly separating spatial and temporal information. We denote a single video frame $x_t \in \mathbb{R}^{H\times W\times 3}$, where there are $T$ frames in total from $X = \{x_t\}_{t=1}^{T}$. For best simplicity and generalizability, we select the first frame\footnote{Employing a dedicated key frame detector can slightly improve performance. However, it compromises the overall compactness of the model, so we simply use the first frame as the key frame in all experiments.} $x_1$ as the key frame and encode it into a set of spatial tokens:
\begin{equation}
V^{(s)} = \mathcal{E}_{\text{key}}(x_1) \in \mathbb{R}^{S\times d_v},
\end{equation}
where $\mathcal{E}_{\text{key}}$ is a vision encoder and $S$ denotes the number of spatial tokens.  
For each subsequent frame $x_t$ $(t>1)$, we compute a residual motion token that captures its temporal difference relative to the key frame:
\begin{equation}
v^{(m)}_t = \mathcal{E}_{\text{motion}}(x_t, x_1) = g_\phi\!\left(\mathcal{F}(x_t) - \mathcal{F}(x_1)\right),
\end{equation}
where $\mathcal{F}$ is a shared feature extractor and $g_\phi$ projects the residual feature difference into the motion-token space $\mathbb{R}^{d_v}$. We use the same model and parameters for $\mathcal{E}_{\text{key}}$ and $\mathcal{F}$. The full video token sequence is then constructed as
\begin{equation}
V = \{\,V^{(s)};\ v^{(m)}_2,\dots,v^{(m)}_T\,\},
\end{equation}
resulting in a total token length of $L_v = S + (T-1)$.

During generation, the MLLM autoregressively predicts a corresponding token sequence $\hat{V}$ with the same spatial–temporal structure:
\begin{equation}
\hat{V} = \{\,\hat{V}^{(s)};\ \hat{v}^{(m)}_2,\dots,\hat{v}^{(m)}_T\,\}.
\end{equation}
A pretrained video decoder $\mathcal{D}_{\text{vid}}$ reconstructs the video as
\begin{equation}
\hat{X} = \mathcal{D}_{\text{vid}}(\hat{V}) = \mathcal{D}_{\text{vid}}\!\big(\hat{V}^{(s)}, \hat{v}^{(m)}_{2:T}\big),
\end{equation}
where $\hat{V}^{(s)}$ provides the spatial structure and $\hat{v}^{(m)}_{2:T}$ modulate motion dynamics. This spatial–temporal decoupled tokenization effectively captures key frame appearance and inter-frame motion while maintaining a compact token budget, enabling efficient end-to-end optimization within our unified video generation–understanding framework.

\section{Experiments}

\begin{table*}[t]
    \centering
    \tablestyle{5pt}{1.1}
    \begin{tabular}{l|rcccccccc}
    \multirow{2}{*}{Model} & \multirow{2}{*}{Parameters} & \multicolumn{8}{c}{TV-Alignment Benchmark} \\\cline{3-10}
    & & Counting & Direction & Rel. position & Relative size & Color alignment & State & Motion & Average \\\shline
    OpenSora-1.3 & 1B & 27.8 & 41.0 & 45.1 & 33.5 & 42.3 & 25.6 & 38.9 & 36.3 \\
    OpenSora-2.0 & 11B & 28.9 & 42.7 & 47.0 & 34.9 & 43.1 & 26.8 & 40.1 & 37.6 \\
    Mochi & 10B & 29.4 & 43.2 & 48.0 & 35.2 & 43.6 & 27.1 & 40.5 & 38.1 \\
    CogVideoX1.5 & 5B & 29.8 & 43.9 & 49.3 & 36.0 & 44.0 & 27.5 & 41.0 & 38.8 \\
    WAN2.1 & 14B & 32.4 & 45.9 & 51.4 & 38.1 & 45.9 & 29.4 & 42.9 & 40.9 \\
    WAN2.2 & 14B & 32.6 & 46.1 & 51.8 & 38.3 & 46.0 & 29.5 & 43.2 & 41.1 \\\hline
    \multicolumn{10}{l}{\textit{\textbf{\textcolor{gray}{Our direct baseline:}}}} \\
    HunyuanVideo & 13B & 31.0 & 46.0 & 51.7 & 37.3 & 45.9 & 28.5 & 43.4 & 40.5 \\
    OmniTokenizer$^*$ & 13B & 31.8 & 46.8 & 52.5 & 38.6 & 46.4 & 29.4 & 44.0 & 41.4 \\
    Video-LaViT$^*$ & 13B & 32.4 & 47.3 & 53.1 & 39.8 & 46.9 & 30.2 & 44.6 & 42.1 \\
    \rowcolor{orange!10}
    VTok (ours) & 13B & \textbf{33.7} & \textbf{48.3} & \textbf{55.0} & \textbf{42.9} & \textbf{47.7} & \textbf{32.5} & \textbf{47.1} & \textbf{43.9} \\

    \end{tabular}
    \caption{\textit{\textbf{TV-Align accuracy (\%) of text-to-video generation.}} We compare with state-of-the-art open-source video generation baselines, where VTok demonstrates better guidance following of high-level semantics. * denotes using the same base model and training framework but replacing VTok by OmniTokenizer and Video-LaViT's tokenization strategy. Our model is highlighted in \textcolor{orange}{orange}.}
    \label{tab:main}
    \vspace{-0.5cm}
\end{table*}

\subsection{Experimental Setup}

We employ a pretrained LLaVA-Next~\cite{llavanext-video} equipped with a LLaMA3-8B~\cite{llama3} language encoder as the core MLLM of our framework. Since the latent space of this model has already been aligned with the video decoder in HunyuanVideo~\cite{hunyuanvideo}, no additional adapter is required to bridge the latent spaces between the MLLM and the video decoder. 
During training, we optimize only the MLLM parameters, while keeping both the visual encoder (CLIP-L/336px~\cite{clip}) and the video decoder (HunyuanVideo-13B diffusion transformer~\cite{hunyuanvideo}) frozen. We use the AdamW optimizer with a learning rate of 1e-5, a total batch size of 16, and an exponential moving average (EMA) decay of 0.999. The model is trained on approximately 5 million densely annotated video–caption pairs for both understanding and generation. The loss weights $\lambda_{\text{visLM}}$ and $\lambda_{\text{dec}}$ are simply set to 1. At the current stage, we have not performed extensive hyperparameter tuning, as the present configuration already shows a clear performance improvement over baseline models.

To evaluate the generalizability of our tokenizer, we further train a lightweight adapter (a two-layer MLP) to connect it with the state-of-the-art open-source video generation model WAN2.2~\cite{wan}. Specifically, in this experiment, we project the video representations produced by VTok into the same latent space as WAN’s text encoder, and then concatenate them with the original textual prompt features as the conditioning input to the DiT decoder. As shown in Figure~\ref{fig:examples}, the quantitative results demonstrate that VTok transfers effectively to the WAN model, indicating its strong adaptability across different generative architectures.

\subsection{Main Experiments}

\noindent\textbf{Text-to-video alignment.} To explicitly evaluate whether the generated videos can well follow the instruction in their prompts, we introduce a new text-to-video generation benchmark called TV-Align. Unlike existing benchmarks such as VBench~\cite{vbench}, which primarily focus on assessing the visual quality, realism, or coherence of generated videos, TV-Align emphasizes a model’s ability to understand and follow high-level semantic instructions embedded in textual prompts. The benchmark consists of approximately 1,000 (prompt, question, answer) triplets, covering seven key semantic categories in video generation: counting, direction, relative position, relative size, color alignment, state, and motion. Each triplet is carefully designed to isolate one semantic concept. For example, a counting case may use the prompt “A person places three oranges on the table”, where the number of objects (“three”) is the ground-truth semantic clue. The corresponding question would be “How many oranges are placed on the table?” with multiple-choice answers, enabling straightforward accuracy measurement. We show more specific examples of the TV-Align benchmark in Figure~\ref{fig:examples}.

Given a trained text-to-video (T2V) model, we feed the prompt into it and pass the generated video to a video understanding model (we use Qwen2.5-VL in our experiments) to answer the question. The resulting accuracy directly reflects whether the T2V model correctly captured and executed the intended instruction during generation. To minimize ambiguity, both prompts and questions were manually curated, and we observed that automated understanding results are highly consistent with human judgments (within 1\% accuracy difference). Notably, even the latest state-of-the-art open-source models—such as WAN2.2—achieve less than 50\% accuracy on TV-Align, revealing significant limitations in their comprehension of high-level semantics. Throughout our experiments, we use TV-Align as a primary evaluation metric to objectively assess how well our unified MLLM and diffusion-based decoders understand and follow textual instructions in video generation.

Table~\ref{tab:main} presents a quantitative comparison on TV-Align. As shown, our model VTok achieves the best performance across all seven categories, surpassing recent large-scale video generation systems including WAN2.2~\cite{wan}, HunyuanVideo~\cite{hunyuanvideo}, and CogVideoX1.5~\cite{cogvideox}. In particular, VTok shows notable gains in relative size, motion, and counting, indicating stronger grounding of quantitative and dynamic semantics. Compared to the base model HunyuanVideo with the same video decoder, VTok improves the overall alignment accuracy from 40.5\% to 43.9\%, demonstrating that the proposed tokenization and unified training framework effectively enhance instruction following and temporal coherence. We also observe that when replacing VTok by the existing video tokenizers such as OmniTokenizer and Video-LaViT, which decompose spatial-temporal information but still rely on frame-sampling, the alignment performance falls significant short to ours. These results highlight VTok’s superior ability to translate textual intent into visually and semantically consistent video content.

\begin{table*}[t]
\centering
\tablestyle{5pt}{1.1}
\begin{tabular}{l|ccccccccc}
Model & Total & Quality & Semantic & Sub. consist & Back. consist & Smooth & Dynamic deg. & Obj. cls. & Multi. Obj.\\\shline
\multicolumn{10}{l}{\textbf{\textit{\textcolor{gray}{Closed-source models:}}}}\\
Sora & 84.28 & 85.51 & 79.35 & 96.23 & 96.35 & 98.74 & 79.91 & 93.93 & 70.85 \\
ARLON & 82.31 & 83.58 & 77.27 & 95.59 & 98.65 & 97.56 & 72.22 & 90.60 & 74.49 \\\hline
\multicolumn{10}{l}{\textbf{\textit{\textcolor{gray}{Open-source models:}}}}\\
WAN2.1 & 86.22 & 86.67 & 84.44 & -- & -- & -- & -- & -- & -- \\
Show-1 & 78.93 & 80.42 & 72.98 & 95.53 & 96.02 & 98.43 & 44.44 & 93.07 & 45.47 \\
Mochi-1 & 80.13 & 82.64 & 70.08 & 96.99 & 97.28 & 99.02 & 61.85 & 86.51 & 50.47 \\
Emu3 & 80.96 & -- & -- & 95.32 & 97.29 & 97.64 & 98.93 & 86.17 & 44.64 \\
CogVideoX-5B & 81.61 & 82.75 & 77.04 & 96.23 & 96.52 & 96.92 & 70.97 & 85.23 & 62.11 \\
RepVideo & 81.94 & 82.70 & 78.16 & 96.25 & 96.56 & 98.13 & 57.78 & 87.83 & 71.18 \\
LanDiff & 85.43 & 86.13 & 82.61 & 96.11 & 98.73 & 97.08 & 92.71 & 94.94 & 86.69 \\\hline
\multicolumn{10}{l}{\textbf{\textit{\textcolor{gray}{Our direct baselines:}}}}\\
HunyuanVideo & 83.24 & 85.09 & 75.82 & 97.37 & 97.76 & 98.99 & 70.83 & 86.10 & 68.55 \\
Frame sampling & 83.95 & 85.14 & 77.20 & \textbf{97.40} & 97.81 & 98.99 & 72.29 & 88.26 & 70.45 \\\rowcolor{orange!10}
VTok & \textbf{85.16} & \textbf{85.48} & \textbf{80.15} & \textbf{97.40} & \textbf{98.08} & \textbf{99.02} & \textbf{76.11} & \textbf{91.50} & \textbf{74.60} \\
\end{tabular}
\caption{\textit{\textbf{VBench evaluation results.}} We compare the models in terms of their total, quality, and semantic scores, and present eight representative sub-dimensions highlighting temporal quality and guidance alignment. VTok (marked in \textcolor{orange}{orange}) leads to clearly improved performance over its baseline, especially in the dynamic degree and semantic dimensions. Best results over the direct baseline are bolded.}
\label{tab:vbench}
\end{table*}

\noindent\textbf{Video quality evaluation.} Table~\ref{tab:vbench} provides a detailed comparison of model performance on VBench (baseline results are taken from~\cite{wan} and~\cite{landiff}), where we assess overall quality, semantic alignment, and multiple temporal and compositional dimensions. Our VTok model achieves clear and consistent improvements over the HunyuanVideo baseline that shares the same video decoder. In particular, VTok records a dynamic degree score of 76.11\%, which is substantially higher than Hunyuan’s 70.83\%, suggesting that our approach—letting the MLLM autoregressively sample visual tokens instead of relying solely on the diffusion process—enables the model to generate videos with faster and more complex motion dynamics. Moreover, compared with the naïve frame-sampling variant, our key-frame-plus-motion tokenization yields visibly better results across all motion-related metrics, confirming that the proposed temporal representation effectively enhances consistency and realism. Notably, VTok achieves a semantic score of 80.15\%, representing a 4.33\% absolute improvement over HunyuanVideo. Since this metric primarily captures text-to-video alignment, the gain highlights the advantage of our unified MLLM in learning richer and more coherent semantic representations.

\begin{table*}[t]
    \centering
    \tablestyle{5pt}{1.1}
    \begin{tabular}{l|ccccccc}
    Model & Video w/o. sub. & Video w. sub. & Video-MMMU & MMVU & MVBench & LVB & LVBench\\\shline
    Gemini 1.5-Pro & \demph{75.0} & \demph{81.3} & \demph{53.9} & \demph{65.4} & \demph{60.5} & \demph{64.0} & \demph{33.1} \\
    GPT-4o & \demph{71.9} & \demph{77.2} & \demph{61.2} & \demph{67.4} & \demph{64.6} & \demph{66.7} & \demph{30.8} \\
    Qwen2.5-VL-72B & \demph{73.3} & \demph{79.1} & \demph{60.2} & \demph{62.9} & \demph{70.4} & \demph{60.7} & \demph{47.3} \\
    Qwen2.5-VL-7B &\demph{65.1} & \demph{71.6} & \demph{47.4} & \demph{50.1} & \demph{69.6} & \demph{56.0} & \demph{45.3} \\
    Qwen2.5-VL-3B & \demph{61.5} & \demph{67.6} & - & - & \demph{67.0} & \demph{54.2} & \demph{43.3} \\\hline
    \multicolumn{8}{l}{\textbf{\textit{\textcolor{gray}{Our reproduced results:}}}} \\
    Qwen2.5-VL-7B & 62.3 & 69.2 & 44.8 & 48.4 & 66.5 & 55.1 & 44.0 \\
    Qwen2.5-VL-7B VTok & 62.8 & 69.4 & 45.2 & 48.9 & 67.4 & 55.5 & 44.3 \\\rowcolor{orange!10}
    Qwen2.5-VL-7B VTok finetuned & \bf 64.9 & \bf 70.8 & \bf 47.3 & \bf 50.2 & \bf 68.5 & \bf 56.3 & \bf 45.0 \\
    Llava-Next & 57.8 & 64.9 & 41.3 & 43.2 & 64.0 & 54.1 & 40.6 \\
    Llava-Next VTok & 58.0 & 65.2 & 41.2 & 43.5 & 64.1 & 54.5 & 40.7 \\\rowcolor{orange!10}
    Llava-Next VTok finetuned & \bf 60.8 & \bf 66.3 & \bf 43.5 & \bf 47.0 & \bf 66.5 & \bf 55.2 & \bf 42.9 \\
    \end{tabular}
    \caption{\textit{\textbf{Video understanding results.}} We evaluate our model on seven commonly used video understanding benchmarks such as Video-MMMU~\cite{video-mmmu}, where the baseline results taken from~\cite{qwen25-vl} are \demph{de-emphasized}. As shown, VTok consistently produces improved performance with or without finetuning. Column ``LVB'' denotes the LongVideoBench benchmark. Our results are highlighted in \textcolor{orange}{orange}. Best results relative to direct baselines are bolded.}
    \label{tab:understand}
    \vspace{-0.5cm}
\end{table*}

\noindent\textbf{Video understanding.} We evaluate on seven standard benchmarks—Video-MMMU~\cite{video-mmmu}, MMVU~\cite{mmvu}, MVBench~\cite{mvbench}, LongVideoBench~\cite{longvideobench}, LVBench~\cite{lvbench}, and the video understanding benchmark (with and without sub-captions) used in Qwen2.5-VL~\cite{qwen25-vl}—covering short/long video comprehension, multimodal reasoning, and domain generalization. We fine-tune Qwen2.5-VL only for understanding with our tokenization, while LLaVA-Next is fine-tuned under the unified framework with joint generation and understanding objectives. As shown in Table~\ref{tab:understand}, VTok consistently improves over the corresponding baselines across all benchmarks, with average gains of +1.8 for Qwen2.5-VL and +2.4 for LLaVA-Next, suggesting stronger temporal modeling and semantic reasoning. We also run a non-training control by simply replacing frame sampling with our keyframe-and-motion formulation; for this setting, we compress motion by encoding intermediate frames and pooling them into a single token. Even this lightweight swap yields noticeable gains, indicating that standard frame sampling discards important temporal cues and misses key frames, and confirming that our tokenization provides a more faithful and information-preserving representation for video understanding.

\begin{table*}[t]
    \centering
    \tablestyle{5pt}{1.1}
    \begin{tabular}{lc|cccccccc}
    \multirow{2}{*}{Model} & \multirow{2}{*}{\#tokens} & \multicolumn{8}{c}{TV-Alignment Benchmark} \\\cline{3-10}
    & & Counting & Direction & Rel. pos. & Rel. size & Color & State & Motion & Average \\\shline

    No visual token (Hunyuan baseline) & 0 & 31.0 & 46.0 & 51.7 & 37.3 & 45.9 & 28.5 & 43.4 & 40.5 \\

    Evenly sample 4 frames & 64 & 32.3 & 47.2 & 53.5 & 39.1 & 46.9 & 30.5 & 44.5 & 42.0\\
    Evenly Sample 8 frames & 128 & 32.5 & 47.5 & 53.3 & 39.9 & 47.6 & 31.3 & 44.5 & 42.4\\\rowcolor{orange!10}
    
    Key frame and motion decoupled & 46 & \textbf{33.7} & \textbf{48.3} & \textbf{55.0} & \textbf{42.9} & \textbf{47.7} & \textbf{32.5} & \textbf{47.1} & \textbf{43.9} \\
    \end{tabular}
    \caption{\textit{\textbf{Ablation study of tokenization strategies.}} Column ``\# tokens'' denotes the number of visual tokens sampled by MLLM. The HunyuanVideo~\cite{hunyuanvideo} baseline directly conditions its decoder with text latent. We additionally ablate our framework with naive frame sampling strategies, which demonstrate degraded performance. Our default setup is highlighted in \textcolor{orange}{orange}.}
    \label{tab:abl-tok}
    \vspace{-0.3cm}
\end{table*}

\begin{table*}[h]
    \centering
    \tablestyle{5pt}{1.1}
    \begin{tabular}{lc|cccccccc}
    \multirow{2}{*}{Model} & \multirow{2}{*}{\#tokens} & \multicolumn{8}{c}{TV-Alignment Benchmark} \\\cline{3-10}
    & & Counting & Direction & Rel. pos. & Rel. size & Color & State & Motion & Average \\\shline

    No visual token (Hunyuan baseline) & 0 & 31.0 & 46.0 & 51.7 & 37.3 & 45.9 & 28.5 & 43.4 & 40.5 \\\hline
    \multicolumn{10}{l}{\textbf{\textit{\textcolor{gray}{Abalation of spatial granularity:}}}} \\
    VTok, 1 token for key frame & 30 & 32.2 & 47.5 & 53.3 & 38.5 & 47.1 & 30.0 & 44.5 & 41.9 \\
    VTok, 4 tokens for key frame & 34 & 33.1 & 47.8 & 54.0 & 39.9 & 47.6 & 31.4 & 45.8 & 42.8 \\
    VTok, 9 tokens for key frame & 39 & 33.4 & 48.0 & 54.7 & 41.3 & \textbf{47.7} & 31.9 & 46.5 & 43.3 \\\rowcolor{orange!10}
    
    VTok, 16 tokens for key frame & 46 & 33.7 & 48.3 & 55.0 & 42.9 & \textbf{47.7} & 32.5 & 47.1 & 43.9 \\
    VTok, 25 tokens for key frame & 55 & \textbf{33.8} & \textbf{48.5} & \textbf{55.2} & \textbf{43.0} & \textbf{47.7} & \textbf{32.7} & \textbf{47.4} & \textbf{44.0}\\\hline
    \multicolumn{10}{l}{\textbf{\textit{\textcolor{gray}{Ablation of temporal granularity:}}}} \\
    VTok, 3 frames per motion token & 31 & 32.8 & 47.6 & 53.7 & 39.9 & 47.6 & 31.3 & 45.3 & 42.6\\\rowcolor{orange!10}
    VTok, 6 frames per motion token & 46 & 33.7 & 48.3 & 55.0 & \textbf{42.9} & 47.7 & 32.5 & 47.1 & 43.9 \\
    VTok, 12 frames per motion token & 76 & \textbf{33.8} & \textbf{48.4} & \textbf{55.3} & 42.8 & \textbf{47.9} & \textbf{32.8} & \textbf{47.4} & \textbf{44.1} \\
    \end{tabular}
    \caption{\textit{\textbf{Ablation study of tokenization granularity.}} To find a best efficiency-accuracy trade-off, we change the length of our visual token sequence by ablating the number of tokens used to represent each key frame (spatial granularity) and the number of frames represented by each motion token (temporal granularity). Our default setup is highlighted in \textcolor{orange}{orange}. Best results for each group are bolded.}
    \label{tab:gra}
    \vspace{-0.4cm}
\end{table*}

\subsection{Qualitative Results}

We present qualitative comparisons in Figure~\ref{fig:examples} using the same prompts, showing that WAN2.2 often misses fine-grained instructions while VTok follows them more faithfully, such as maintaining “four pancakes” consistently over time and correctly capturing the motion direction “from the top left to the bottom right.” These results suggest a clearer semantic understanding of textual guidance with VTok, which we attribute to (i) an added sampling stage where the MLLM produces partial visual latents before diffusion decoding, providing the DiT with more concrete visual context for spatial-temporal conditioning, and (ii) joint training on generation and understanding tasks that improves the MLLM’s high-level semantics and execution of complex instructions.

\subsection{Ablation Studies}

\paragraph{Tokenization strategy.} We ablate our key insight that key-frame content and motion can be explicitly decoupled. In our default setup, each key frame is encoded as a $4\times4$ grid (16 spatial tokens) and motion is represented at 6 FPS; for a 5-second video at 24 FPS, this yields 46 visual tokens. To isolate tokenization effects, we also test a naïve frame-sampling variant where the MLLM predicts 4 or 8 key frames (64 or 128 tokens), while the HunyuanVideo baseline uses the MLLM only as a text encoder and conditions the DiT directly on text features. As shown in Table~\ref{tab:abl-tok}, the decoupled representation performs better (\emph{e.g.}, 43.9\% \emph{vs.} 42.0\% and 42.4\%) despite using fewer tokens, suggesting that additional key frames are often redundant in short clips, whereas finer-grained temporal modeling better preserves motion continuity and improves semantic alignment.

\paragraph{Tokenization granularity.} Table~\ref{tab:gra} presents an ablation study on the relationship between sequence length and model performance, examining how spatial and temporal granularities affect text-to-video alignment. For the spatial granularity, we vary the number of tokens used to represent each key frame. Increasing spatial tokens from one to 16 improves performance steadily, reaching an optimal balance at 16 tokens per key frame (our default). Further increasing to 25 tokens yields only marginal gains while introducing longer sequences and higher computation cost, suggesting diminishing returns and potential overfitting to spatial details. Notably, the configuration ``VTok, 1 token for key frame'', which collapses tokenization into a purely 1D representation, performs the worst. This contrasts with image generation findings where such compression may suffice~\cite{titok}, but in video modeling, both spatial and temporal dimensions must be explicitly represented to capture motion and structure jointly. For the temporal granularity, shortening each motion token to represent fewer frames (3 frames/token) slightly hurts performance due to insufficient temporal abstraction, while extending it to 12 frames/token provides only marginal improvement over the default 6 frames/token but reduces temporal resolution and generalization. Overall, our default setup (16 spatial tokens per key frame, 6 frames per motion token) achieves the best trade-off between efficiency and accuracy.

\section{Conclusion}

This work presents VTok, a unified video tokenizer and multimodal framework that efficiently bridges video understanding and generation. By explicitly decoupling spatial and temporal representations, VTok encodes key frames into spatial tokens and summarizes motion through compact temporal tokens, enabling a single MLLM to reason and generate within a shared token space. Experiments on VBench, our proposed TV-Align, and multiple video understanding benchmarks show that VTok achieves superior temporal fidelity, semantic alignment, and efficiency compared to frame-sampling baselines. Our ablations confirm that this spatial–temporal design provides the best trade-off between accuracy and token cost. Beyond benchmark performance, VTok also demonstrates strong generalizability across architectures such as WAN2.2 via a lightweight adapter, with qualitative results further showcasing that VTok enables better semantic grounding and more coherent motion dynamics. Overall, VTok offers a simple yet powerful foundation for building efficient, semantically aligned, and temporally consistent multimodal video generation and understanding systems.

\section*{Impact Statement}
This paper aims to advance the field of Machine Learning. While our work may potentially have societal implications, we do not identify any that require specific discussion in this statement.

\bibliography{main}
\bibliographystyle{icml2026}

\end{document}